\documentclass[utf8]{frontiersSCNS} 
\usepackage{url}
\usepackage{microtype,subcaption}
\usepackage[onehalfspacing]{setspace}
\usepackage{tikz}
\usepackage{stfloats}
\usepackage{multicol}
\usepackage{multirow}
\usepackage{rotating}
\usepackage{mathtools}
\usepackage{soul}
\DeclareRobustCommand{\ballnumber}[1]{\tikz[baseline=(myanchor.base)] \node[circle,fill=.,inner sep=1pt] (myanchor) {\color{-.}\bfseries\footnotesize #1};}
\DeclareRobustCommand\boxnum[1]{\tikz[baseline=(char.base)]{\node[shape=rectangle,draw,inner sep=2pt] (char) {#1};}}
\usepackage{framed}
\usepackage{mathptmx} \usepackage{times} \usepackage{amsmath} \usepackage{amssymb}  
\usepackage[]{xcolor}
\usepackage{dcolumn}
\newcolumntype{d}[1]{D{.}{.}{#1}}
\usepackage[normalem]{ulem}
\usepackage{siunitx}            
 \DeclareCaptionSubType{figure}   
\usepackage{cleveref}
\newcommand{\mrka}[1]{\textcolor{black}{#1}} 
\newcommand{\mrkb}[1]{\textcolor{black}{#1}} 
\def\keyFont{\fontsize{8}{11}\helveticabold }
\def\firstAuthorLast{Mo {et~al.}} \def\Authors{An Mo\,$^{1}$, Fabio Izzi\,$^{1,2}$ Daniel F. B. Haeufle\,$^{2}$ and  Alexander Badri-Spr\"owitz$^{1,*}$}
\def\Address{$^{1}$Dynamic Locomotion Group, Max Planck Institute for Intelligent Systems, Stuttgart, Germany \\
$^{2}$Hertie-Institute for Clinical Brain Research, University of T\"ubingen, T\"ubingen, Germany}
\def\corrAuthor{Alexander Badri-Spr\"owitz}
\def\corrEmail{sprowitz@is.mpg.de}

\begin{document}
\onecolumn
\firstpage{1}
\title[Effective damping in legged locomotion]{Effective Viscous Damping Enables Morphological Computation in Legged Locomotion} 
\author[\firstAuthorLast ]{\Authors} \address{} \correspondance{} 
\extraAuth{}
\maketitle
\begin{abstract} 
Muscle models and animal observations suggest that physical damping is beneficial for stabilization. Still, only a few implementations of physical damping exist in compliant robotic legged locomotion. It remains unclear how physical damping can be exploited for locomotion tasks, while its advantages as sensor-free, adaptive force- and negative work-producing actuators are promising. In a simplified numerical leg model, we studied the energy dissipation from viscous and Coulomb damping during vertical drops with ground-level perturbations. A parallel spring- damper is engaged between touch-down and mid-stance, and its damper auto-decouples from mid-stance to takeoff. Our simulations indicate that an adjustable and viscous damper is desired. In hardware we explored effective viscous damping and adjustability, and quantified the dissipated energy. We tested two mechanical, leg-mounted damping mechanisms: a commercial hydraulic damper, and a custom-made pneumatic damper. The pneumatic damper exploits a rolling diaphragm with an adjustable orifice, minimizing Coulomb damping effects while permitting adjustable resistance. Experimental results show that the leg-mounted, hydraulic damper exhibits the most effective viscous damping. Adjusting the orifice setting did not result in substantial changes of dissipated energy per drop, unlike adjusting the damping parameters in the numerical model. Consequently, we also emphasize the importance of characterizing physical dampers during real legged impacts to evaluate their effectiveness for compliant legged locomotion.
\tiny
 \keyFont{ \section{Keywords:} damping, energy dissipation, legged locomotion, ground disturbance, drop test, rolling diaphragm} \end{abstract}

\section{Introduction}
\begin{figure*}[t]
\centering
\includegraphics[width=0.95\textwidth]{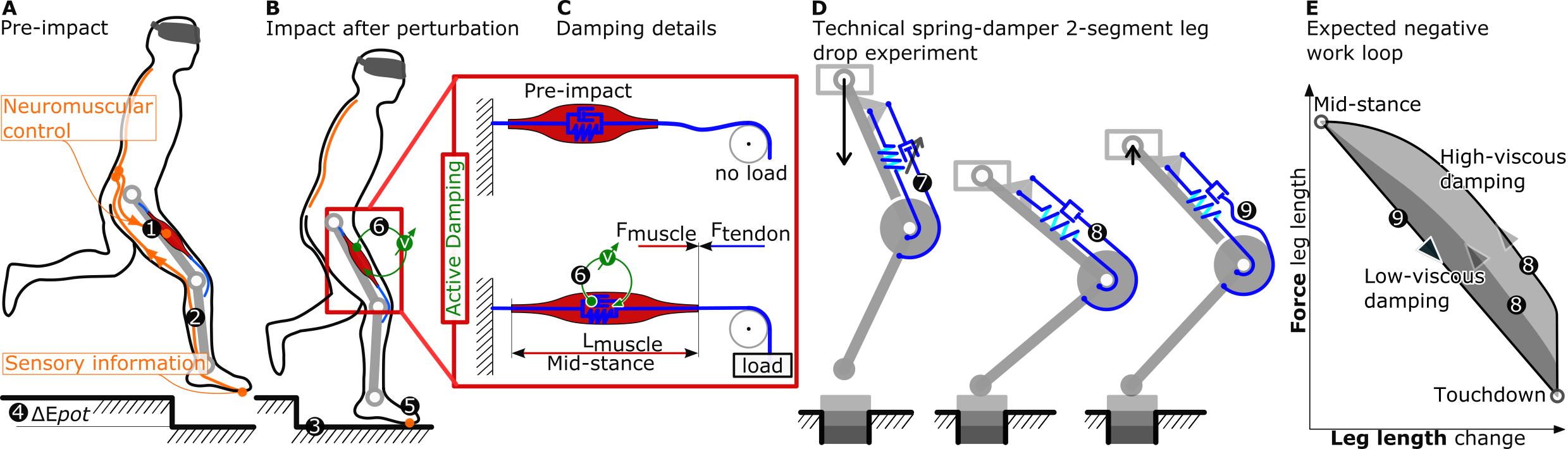}
\caption{\textbf{(A-C) Problem identification, and related research question.} The limited \textcolor{orange}{nerve conduction velocity in organic tissue} \citep{more_scaling_2010} \ballnumber{2} presents a significant hazard in legged locomotion. \textcolor{OliveGreen}{Local neuromuscular strategies} \ballnumber{6} provide an \textcolor{OliveGreen}{alternative means} of \textbf{timely and tunable force and power production}. Actuators like the indicated \textcolor{red}{knee extensor muscle} keep the leg extended during stance phase (muscle length L\textsubscript{muscle}) by producing the appropriate amount of muscle force (F\textsubscript{muscle}), correctly \textbf{timed}. \textcolor{orange}{Neuromuscular control} \ballnumber{1} plays a major role in initiating and producing these active muscle forces, but works best only during unperturbed locomotion. \textcolor{orange}{Sensor information} from foot contact travels via \textcolor{orange}{nerves bundles} \ballnumber{2} to the spinal cord, but with \textbf{significant time delays} in the range of $t=\SI{40}{ms}$ \citep[for \SI{1}{m} leg length]{more_scaling_2018} and more. Hence, the locomotion control system can become \textbf{`sensor blind' due to conduction delays, for half a stance phase}, and can miss unexpected perturbations like the depicted step-down. During step-down perturbations \ballnumber{3} additional energy \ballnumber{4} is inserted into the system. Viscous damper-like mechanisms \textbf{produce velocity dependent counter-forces, and can dissipate kinetic energy}. \textcolor{OliveGreen}{Local neuromuscular strategies} \ballnumber{6} producing tunable, viscous damping forces would \textbf{act instantaneously and adaptively}. Such strategies \ballnumber{6} could also be robust to uncontrolled and harsh impacts of the foot after perturbations \ballnumber{5}, better than \textcolor{orange}{sensor-based strategies.} In this work \textbf{(D)}, we are testing and characterizing spring-damper configurations mounted to a two-segment leg structure, during rapid- and slow-drop experiments, for their feasibility to \textcolor{OliveGreen}{physically} and instantaneously produce tunable, speed-dependent forces extending the leg. Work loops \textbf{(E)} will indicate how much effective negative work is dissipated, between touch-down and mid-stance. Prior to impact \ballnumber{7} and during the leg loading \ballnumber{8} the spring-damper's tendons act equally. \mrka{Starting at mid-stance, the main spring extends the knee, leading to leg extension and leaving the damper's tendon slack \ballnumber{9}}.}
\label{fig:bigPicture}
\end{figure*}

While less understood, damping likely plays an essential role in animal legged locomotion. Intrinsic damping forces can potentially increase the \mrka{effective} force output during unexpected impacts \citep{muller_kinetic_2014}, reduce control effort \citep{Haeufle2014a}, stabilize movements \citep{Seipel_2012,Secer2013,Abraham2015}, and reject unexpected perturbations \citep{Haeufle2010a,Kalveram2012}, e.g., sudden variations in the ground level (\Cref{fig:bigPicture}). Stiffness, in comparison, has been studied extensively in legged locomotion. Its benefits have been shown both in numerical simulations, e.g., through spring-loaded inverted pendulum (SLIP) models \citep{mochon1980ballistic,Blickhan2007}, and physical springy leg implementations \citep{Hutter2016,sprowitz2013towards,ruppert_series_2019}.

What combines both mechanical stiffness and intrinsic, mechanical damping is their sensor- and computational-free action. A spring-loaded leg joint starts building up forces exactly at the moment of impact. Mechanical stiffness, or damping, acts instantaneously, and are not subject to delays from post-processing sensor data \citep{grimminger_open_2020}, delays from limited nerve conductive velocities \citep{more_scaling_2018}, or uncertainties in the estimation of the exact timing of swing-to-stance switching \citep{bledt_contact_2018}.

Legged robots commonly exploit \emph{virtual damping}: actively produced and sensory-controlled negative work in the actuators \citep{Seok2015,Hutter2012,Havoutis2013,Kalouche_2007,grimminger_open_2020}. Virtual damping requires high-frequency force control, and actuators mechanically and electrically capable of absorbing peaks in negative work. In comparison, mechanical damping based systems \citep{Hu2019,Garcia2011} act instantaneously, share impact loads with the actuator when in parallel configuration, and require no sensors or control feedback. 
\mrkb{The instantaneous mechanical response of a damper is especially relevant in biological systems, where the neuronal delay may be as large as \SIrange{5}{40}{\%} of the duration of a stance phase \citep{more_scaling_2010}. In such a short time-window, physical damping could help to reject the perturbation \citep{Haeufle2010a,Kalveram2012} by morphological computation, as it mechanically contributes to the rejection of the perturbation, a contribution that otherwise would need to be achieved by a (fast) controller \citep{Zahedi2013,Ghazi-Zahedi2016}.} Hence, physical damping has the potential to contribute to the \emph{morphological computation} \citep{Zahedi2013,Ghazi-Zahedi2016} of a legged system.

\mrka{Compared to virtual damping with proprioceptive sensing strategies \citep{grimminger_open_2020}, a legged robot with physical damping requires additional mechanical components, \mrkb{e.g.}, a fluidic cylinder, and the mechanics to convert linear motion to rotary output. In a cyclic locomotion task, the energy removed by any damper must also be replenished. Hence, from a naive energetic perspective, both virtual and physical damping systems are costly.}

\mrka{Energy dissipation in the form of negative work has been quantified in running birds, and identified as a potential strategy to ‘... reduce the likelihood of a catastrophic fall.’ \cite[p.\,185]{daley_running_2006b}. In virtual point-based control strategies for bipedal running, positive work is inserted into hip joints, and negative work is then dissipated in equal amounts in the spring-damper leg \citep{drama_trunk_2020}. In sum, either physical damping or virtual damping allows removing energy from a legged locomotion system. In this work, we focus on physical damping produced by a viscous damper. We aim towards an understanding of how physical damping can be exploited in legged locomotion and which requirements a damper must fulfill.}

\mrka{We consider two damping principles: viscous damping and Coulomb damping. Viscous damping reacts to a system motion with a force that is linearly (or non-linearly) proportional to its relative acting speed. Coulomb damping generates a constant force, largely independent from its speed \citep{Serafin_2004}.}
\mrkb{From a control perspective, viscous damping can be beneficial for the negotiation of perturbations in locomotion as it approximates the characteristics of a differential, velocity-dependent term. Yet it is unknown how this intuition transfers into reality, where impact dynamics and non-linearities of the leg geometry alter the stance-phase dynamics of locomotion.}

\mrka{Damping in legged locomotion can have other purposes, besides dissipating energy. The authors of \cite[p.\,7]{werner_generation_2017} introduced a damping matrix in the control scheme, which reduced unwanted oscillations in the presence of modeling errors. \cite{tsagarakis2013asymmetric} mount compliant elements with some damping characteristics, which also could reduce oscillations of the system's springy components.}

\begin{figure}[ht]
\centering
\includegraphics[scale = 1]{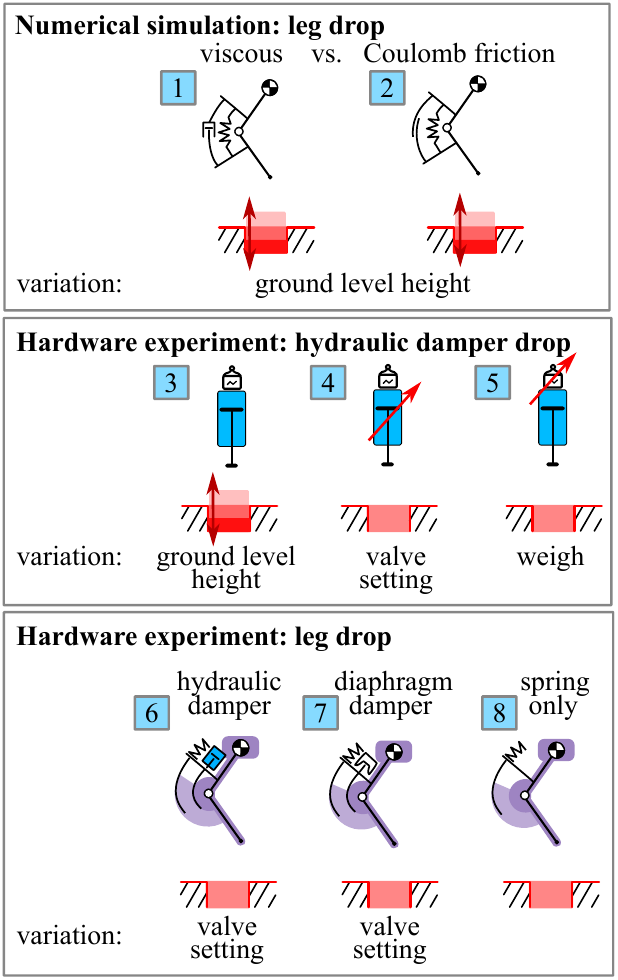}
\caption{Overview: We study the \emph{effective dissipated energy} $E_\mathrm{effective}$ in drop experiments, i.e., the energy dissipation within one drop cycle between touch-down and lift-off (\Cref{fig:highSpeedSnapshots}). We focus on a system design with a damper and a spring, both acting in parallel on the knee joint (\Cref{fig:bigPicture}\textbf{E} and \Cref{fig:legModeling}). No active motor is considered as it is not relevant for the drop scenario, but required for continuous hopping. In \textbf{numerical simulations}, we quantify the difference in energy dissipation between viscous\,\boxnum{1} and Coulomb damping for varying ground level heights\,\boxnum{2} (\Cref{Sec:simulation} and \Cref{fig:energySimulation}). The \textbf{first set of hardware experiments} characterizes the industrial hydraulic damper. For this, we \textbf{drop the isolated damper} (damper only, not mounted in the \mrkb{leg}) on a force sensor and calculate the energy dissipation. We vary the ground level height\,\boxnum{3}, the valve setting\,\boxnum{4} and the drop mass\,\boxnum{5}, to investigate its dynamic characteristics (\Cref{subsec:Results_IsolatedDamper} and \Cref{fig:damper_eval}). For the \textbf{second set of hardware experiments, we drop a 2-segment leg with dampers mounted in parallel to knee springs}. We investigate the energy dissipation dynamics of the hydraulic\,\boxnum{6} and diaphragm damper\,\boxnum{7} by comparing it to a spring-only condition\,\boxnum{8}, where the damper cable is simply detached (\Cref{subsec:Results_Composition_energy_diss} and \Cref{fig:Fl}). We also vary the valve setting on the dampers to test the dynamic adjustability of damping (\Cref{subsec:Results_Adjustability} and \Cref{fig:tunability}).
\label{fig:studydesign}}
\end{figure}

\mrka{In this project, we focus our investigation on the effect of damping during the touch-down (impact) and mid-stance. We chose this simpler drop-down scenario as it captures the characteristics of roughly half a locomotion cycle. A complete cycle would require an active push off phase, and the leg's swing dynamics. Hence, we study the effectiveness of physical damping on the leg's energy dissipation within one drop (touch-down to lift-off), by quantifying its \textbf{effective dissipated energy} $E_\mathrm{effective}$.} We combine insights from numerical simulations and hardware experiments (\Cref{fig:studydesign}). By studying the response of two damping strategies (viscous and Coulomb damping) in numerical drop-down simulations, we investigate how physical damping can influence the dynamics of the impact phase. We then examine how these theoretical predictions relate to hardware experiments with two functionally different, physical dampers.  Hence we explore and characterize the physical damper implementations in a robot leg for their effectiveness in drop-impacts.

\section{Numerical simulation}\label{Sec:simulation}
\begin{figure*}[ht]
\centering
\subcaptionbox{Leg model\label{fig:Model_Simulation}}
{\includegraphics[scale=1.1]{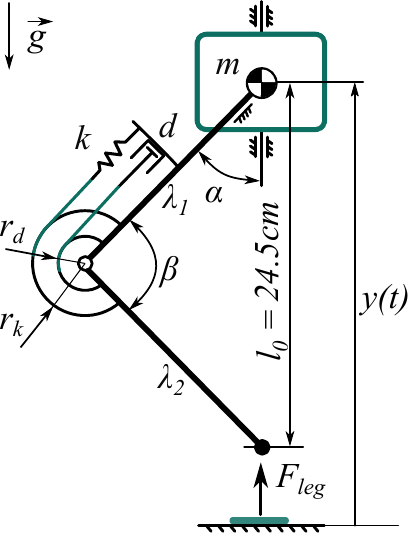}}
\hfill
\subcaptionbox{Leg design\label{subfig:legDesign}}
{\includegraphics[scale=1]{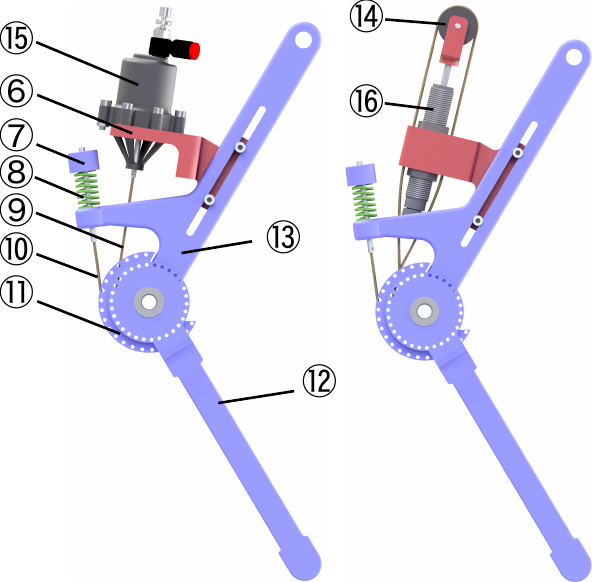}}
\hfill
\subcaptionbox{Drop test bench\label{subfig:legPhoto}}
{\includegraphics[width = 0.24\linewidth]{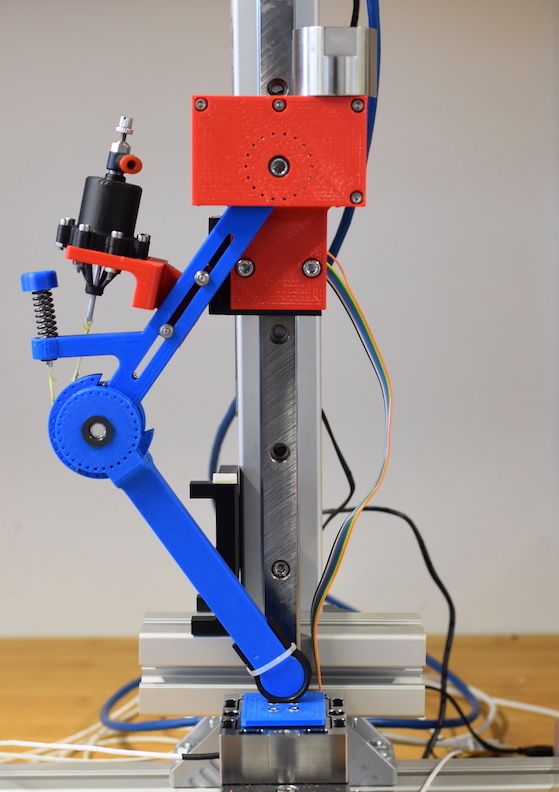}}
\hfill
\caption{\textbf{(a)} 2-segment spring-damper-loaded leg model used for simulation.
\textbf{(b)} Mechanical design of the 2-segment leg. The 
knee pulley{\large \textcircled{\footnotesize 11}} is fixed with the 
lower segment{\large \textcircled{\footnotesize 12}}, coupled with the 
spring{\large \textcircled{\footnotesize 8}} and the 
diaphragm damper{\large \textcircled{\footnotesize 15}} or 
hydraulic damper{\large \textcircled{\footnotesize 16}} via 
cables{\large \textcircled{\footnotesize 9}}{\large \textcircled{\footnotesize 10}}.
\textbf{(c)} Drop test bench with the 2-segment leg.
}
\label{fig:legModeling}
\end{figure*}

We use numerical simulations to theoretically investigate the energy dissipation in a leg drop scenario (\Cref{fig:studydesign}). 
\mrkb{In analogy to our hardware experiment (\Cref{sec:articulatedLeg}), a 2-segment leg with a damper and a spring in parallel on the knee joint is dropped vertically (\Cref{fig:Model_Simulation}).  Once in contact with the ground, the knee flexes and energy is dissipated.} We compare viscous vs. Coulomb damping to investigate which of these two theoretical damping strategies may be more suited for the rejection of ground-level perturbations. Also, we investigate how the adjustment of the damping characteristics influences the dissipated energy. 
\mrkb{In all the damping scenarios investigated, the system is not energy conservative. As we investigate the potential benefit of damping in the initial phase of the ground contact, i.e., from touch-down to mid-stance, we do not consider any actuation. Without actuation or control, the model's dissipated energy is not refilled, unlike in, for example, periodic hopping \citep{Kalveram2012}.}

\vspace{1cm}
\subsection{Model}
\mrkb{The numerical model is a modified version of the 2-segment leg proposed in \cite{Rummel} with an additional damper mounted in parallel to the knee-spring.} The equation describing our leg dynamics is:
\begin{equation}
\label{eqn1}
\ddot{y}\left(t\right) = \frac{F_{leg}\left(t\right)}{m} - g \end{equation}
where $g$ is the gravitational acceleration, $m$ is the leg mass (lumped at the hip), and $y\left(t\right)$ is the time-dependent vertical position from the ground. $F_{leg}\left(t\right)$ is the force transmitted to the hip mass - and the ground - through the leg structure. As such, the force depends on the current phase of the hopping cycle:
\begin{equation}
\label{eq:Fleg}
F_{leg}\left(t\right)= \begin{cases}
0 & \quad \text{, flight phase: } y\left(t\right) > l_0  \\
\displaystyle\frac{y\left(t\right)}{\lambda_{1} \lambda_{2}} \frac{\tau\left(t\right)}{\sin \left(\beta\left(t\right)\right)} &\quad \text{, ground contact: } y\left(t\right) \leq l_0  \end{cases} 
\end{equation}
with segment length $\lambda_i$ and knee angle $\beta\left(t\right)$ (\Cref{fig:Model_Simulation}), $l_0$ is the leg length at impact.
$\tau\left(t\right)$ is the knee torque which is produced by the parallel spring-damper element, as in
\begin{equation}
\tau\left(t\right)= k \, r_k^2 \left(\beta\left(t\right) - \beta_0\right) + \tau_{d}\left(t\right) \label{eqn3}
\end{equation}
with $k$ and $r_k$ being the spring stiffness coefficient and lever arm, respectively. $\tau_{d}\left(t\right)$ is the damping torque, which is set to zero during leg extension, i.e., the damper is only active from impact to mid-stance:
\begin{equation}
\tau_d\left(t\right) = 0 \qquad \text{ if } \dot{\beta}\left(t\right) > 0
\label{eqn4}
\end{equation}
\mrkb{The modeled damper becomes inactive during leg extension, in accordance to our hardware: the tested physical dampers apply forces to the knee's cam via a tendon (\Cref{fig:bigPicture}d, \ballnumber{9}), and this tendon auto-decouples during leg extension.}
By choosing different definitions of the damper torque $\tau_d\left(t\right)$, we can analyse different damper concepts. The model parameters are listed in \Cref{tab:designPara}.

Simulations were performed using MATLAB (the MathWorks, Natick, MA) with ODE45 solver (absolute and relative tolerance of \num{e-5}, max step size of \SI{e-5}{s}). When searching for appropriate settings of the numerical solver, we progressively reduced error tolerances and the maximum step size until convergence of the simulation results in \Cref{tab:energy} to the first non-significant digit.

\begin{table}[bh]
\fontsize{9}{11}\selectfont
\caption{Simulation and hardware parameters}
\label{tab:designPara}
\begin{center}
\begin{tabular}{lccc}
Parameters & Symbol & Value & Unit\\
\hline
Mass                  & $m$       & 0.408   & kg\\
Reference drop height & $h_{0}$ & 14      & cm\\
Spring stiffness      & $k$       & 5900    & N/m\\
Leg segment length    & $\lambda_{1},\lambda_{2}$ & 15 & cm\\
Leg resting length 		& $l_0$ 		&	24.6	& cm\\
Knee resting angle & $\beta_{0}$& 110   & deg\\
Spring lever arm  & $r_{k}$   & 2.5    & cm\\
Damper lever arm  & $r_{d}$   & 2    & cm
\end{tabular}
\end{center}
\end{table}

\subsection{Damping characteristics}
We compared two damping concepts in our numerical simulation: (1) \textit{pure} Coulomb damping, i.e., a constant resistance only dependent on motion direction, and \textit{pure} viscous damping, i.e., a damper torque linearly dependent on the knee angular velocity. Accordingly, we tested two different definitions of $\tau_d$:
\begin{equation}
\tau_{d}\left(t\right)= \begin{cases}
 - d_{c} \, r_d \, \operatorname{sign} (\dot{\beta}\left(t\right)) & \quad \text{, pure Coulomb \mrka{damping}} \\[7pt]
- d_{v} \, r_d^2 \, \dot{\beta}\left(t\right)  &\quad \text{, pure viscous \mrka{damping}}  \end{cases} 
\label{eq:damper_torque_def}
\end{equation}  
where $r_d$ is the damper level arm, $d_{c}$ (in \si{N}) and $d_{v}$ (in \si{Ns/m}) the Coulomb damping and viscous damping coefficients, respectively.
%
\vspace{1cm}
\subsection{Energy dissipation in numerical drop simulations}
With this model, we investigate the difference in energy dissipation in response to step-up/down perturbations (cases \boxnum{1} and \boxnum{2} in \Cref{fig:studydesign}). For each drop test, the numerically modeled leg starts at rest ($\dot{y}\left(t\right)=0$) with a drop height 
\begin{equation} 
h=y(t=0)-l_0
\end{equation}
corresponding to the foot clearance at release. The total energy at release is $E_T \left(h\right)=m\,g\,h$. \mrkb{Given that all model parameters in \Cref{tab:designPara} are fixed, the energy dissipated in a drop becomes a function of the drop height and the damping coefficients: $E_{D}=f_{E_D}(h,d_{c,v})$}.

\mrkb{A simulated drop height $h$ can be seen as a variation 
$\Delta h$ from a reference value $h_0$:
\begin{equation}
    h = h_0 \pm \Delta h
\end{equation}} 
Equal to the hardware experiments, we use $h_{0}=\SI{14}{cm}$ as reference drop height. 
\mrkb{In the reference drop condition, i.e., $h = h_0$, the energy dissipated by damping is $E_{D_0} = E_D\left(h_0\right) = f_{E_D}\left(h_0,d_{c,v}\right)$. $E_{D_0}$ only depends on the damping level, namely the chosen damping strategy (viscous or Coulomb damping) and associated damping coefficient.} We chose five different desired damping levels (set 1-5) as a means of scanning a range in which the damping could be adjusted: for each set, the amount of energy that is dissipated at the reference drop height $E_{D_0}$ differs. 
The chosen $E_{D_0}$ values (\Cref{tab:energy}, column ``Reference height") correspond to proportional levels ($ \left[0.1,0.2,\dotsc,0.5\right] $) of the systems potential energy in terms of the leg resting length $l_0$, as in
\begin{equation}  
E_{D_0} = m\,g\, \left[0.1,0.2,\dotsc,0.5\right]l_0
\end{equation}
\mrkb{This} corresponds to damping configurations that dissipate between $\approx \SI{17}{\%}$ and $\approx \SI{88}{\%}$ of the system's initial potential energy \mrka{at the reference height ($E_{T_0} = E_{T}\left(h_0\right) = m\, g\, h_{0} = \SI{560}{mJ}$)}, as shown in \Cref{tab:energy}, column ``Reference height".
To achieve these desired damping levels, we adjusted the damper parameters $d_c$ and $d_v$ accordingly (\Cref{tab:energy}, column ``Damping coeff."). \mrka{As an example:} for set \num{3}, both damping values were adjusted such that at the reference height $h_0$ both dampers dissipate $E_{D_0} = m\,g\,0.3\,l_0=\SI{295}{mJ}$\mrka{, which corresponds} to \SI{53}{\%} of the total energy $E_{T_0}$.

In the numerical simulations, we focus on the relation between a ground level perturbation $\Delta h$ and the change in energy dissipation -- and their dependency on the damper characteristics. A drop from a height larger than $h_{0}$ corresponds to a step-down ($\Delta h>0$), and a drop from a height smaller than $h_{0}$ to a step-up ($\Delta h<0$). Each condition introduces a change of the total energy of $\Delta E_T=m\,g\,\Delta h$. The change in energy dissipation due to the perturbation is defined as
\begin{equation} 
\Delta E_D\left(\Delta h\right) = E_D\left(h_0 + \Delta h\right) - E_{D_0}
\label{eq:deltaEd}
\end{equation}
which is the difference between the dissipated energy when released from a perturbed height and the dissipated energy when released from the reference height. As a reference, we further define the \emph{full rejection} case where
\begin{equation} 
\label{eq:idealdamping}
\Delta E_{D}\left(\Delta h\right) = \Delta E_T = m\,g\,\Delta h 
\end{equation}
In human hopping a full recovery \mrka{within a single hopping cycle} is not seen during experimental drop down perturbations. Instead, a perturbation of \mrka{$\Delta h = 0.1\,l_0$} is rejected in two to three hopping cycles \citep[\Cref{fig:Fv_height}]{Kalveram2012}. In our results, \mrka{this corresponds to the partial rejections observed with sets \num{2} and \num{3} for $\Delta h = \pm \SI{2.5}{cm}$}.

\begin{table}[tb]
\small
\caption{\textbf{Numerical simulation} Total dissipated energy ($E_D$) in one drop cycle for different drop heights ($h$). \textit{Reference height} is the reference drop height \mrkb{$h = h_0=\SI{14}{cm}$}. During \textit{step up(down)} condition, the drop height is reduced(increased) by $\Delta  h = \SI{2.5}{cm}$. Percentage values indicate the change in dissipated energy ($\Delta E_D$) relative to the change in system total energy ($\Delta E_T$) due to the height perturbations. \mrkb{Each set simulates two separate mechanical dampers (pure viscous or pure Coulomb damping), with damping coefficients chosen to dissipate the same energy at the reference condition, i.e., $E_{D_0}$}. Results of set \num{1}, \num{3} and \num{5} are further described in \Cref{fig:energySimulation}. For all tested conditions, viscous damping outperforms Coulomb damping, as indicated by the always higher percentage values.}
\label{tab:energy}
\centering
\begin{tabular}{cr|c|c||c|c|c}
& & \multicolumn{2}{c||}{\multirow{2}{*}{\textit{Damping coeff.}}} & \textit{Step up} & \textit{Reference height}  & \textit{Step down} \tabularnewline 
& & \multicolumn{2}{c||}{} & \textcolor{black}{$h = h_{0} - \Delta h= 11.5\,cm$}& \textcolor{black}{$h = h_{0}=14\, cm$} & \textcolor{black}{$h = h_{0} + \Delta h= 16.5\,cm$}  \tabularnewline 
& & {$d_v$} & {$d_c$} \vspace{0.1cm} &  $E_{D}$ ($\Delta E_{D}/\Delta E_{T}$) & $E_{D_0}$ ($E_{D_0}/E_{T_0}$) & $E_{D}$ ($\Delta E_{D}/\Delta E_{T}$) \tabularnewline
\hline \rule{0pt}{2ex}
{\multirow{2}{*}{Set 1}} & Viscous & 29.5 Ns/m & 0 N &  \textbf{82 mJ (15\%)} & 97 mJ (17\%) & \textbf{112 mJ (15\%)} \tabularnewline
& Coulomb & 0 Ns/m & 7.7 N & 88 mJ (9\%) & 97 mJ (17\%) & 104 mJ (7\%) \tabularnewline
\hline \rule{0pt}{2ex}
{\multirow{2}{*}{
Set 2}} & Viscous & 68 Ns/m & {0 N} &   \textbf{167 mJ (30\%)} & 197 mJ (35\%) & \textbf{227 mJ (30\%)}  \tabularnewline
& Coulomb &{0 Ns/m} & 17.3 N   &  178 mJ (19\%) & 197 mJ (35\%) & 214 mJ (17\%) \tabularnewline
\hline \rule{0pt}{2ex}
{\multirow{2}{*}{Set 3}} & Viscous & 119.4 Ns/m & {0 N} &  \textbf{249 mJ (46\%)} & 295 mJ (53\%) & \textbf{341 mJ (46\%)} \tabularnewline
& Coulomb & {0 Ns/m} & 29.3 N &  264 mJ (31\%) & 295 mJ (53\%) & 323 mJ (28\%)\tabularnewline
\hline \rule{0pt}{2ex}
{\multirow{2}{*}{Set 4}} & Viscous & 197.1 Ns/m & {0 N} &   \textbf{330 mJ (63\%)} & 393 mJ (70\%) & \textbf{455 mJ (62\%)} \tabularnewline
& Coulomb & {0 Ns/m} & 46.1 N & 346 mJ (47\%) & 393 mJ (70\%) & 436 mJ (43\%) \tabularnewline
\hline \rule{0pt}{2ex}
{\multirow{2}{*}{Set 5}} & Viscous & 349.4 Ns/m & {0 N} &  \textbf{411 mJ (81\%)} & 492 mJ (88\%) & \textbf{572 mJ (80\%)}\tabularnewline
& Coulomb & {0 Ns/m} & 76.3N & 423 mJ (69\%) & 492 mJ (88\%) & 556 mJ (64\%)
\end{tabular}
\end{table}
 
\vspace{1cm}
\subsection{Simulation results}
\begin{figure}[tb]
\setcounter{subfigure}{0}
\centering
\includegraphics[scale = 4]{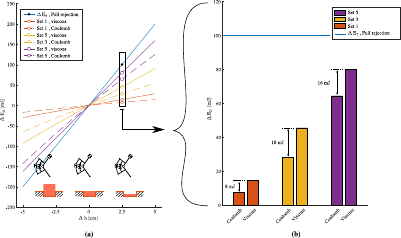}
\caption{\textbf{Numerical simulation} Cases \boxnum{1} and \boxnum{2} from \Cref{fig:studydesign}, \textbf{(a)}: Change of total energy vs. change of drop height for set \num{1}, \num{3} and \num{5}, with damping coefficients as in \Cref{tab:energy}. Continuous lines are viscous damping results, dashed Coulomb damping. Positive perturbations, i.e., $\Delta h > 0$, correspond to step-down perturbations; step-up perturbations, otherwise. The steepest line indicates the slope needed for a \textit{full rejection} of a $\Delta h$ deviation. \mrkb{For each set (1, 3, 5), the damping parameters are matched such that viscous and Coulomb damping dissipate the same energy at the reference height $h_0$ (see \Cref{tab:energy}). Within each set, the viscous damping line is closer to the desired \textit{full rejection} line than the corresponding Coulomb damping line. This means that for the same cost (in the sense of dissipated energy at the reference height) viscous damping always rejects more of ground level perturbation than Coulomb damping}.
\textbf{(b)}: \mrka{$\Delta E_D$} for $\Delta h = \SI{2.5}{cm}$. The horizontal line indicates the amount of energy to dissipate for \textit{full rejection} of $\Delta h$. Energetic advantage of viscous damping over Coulomb damping, as indicated by the spread in the corresponding \mrka{$\Delta E_D$} values, increases from set \num{1} to \num{3}, and reduces from set \num{3} to \num{5}.
\label{fig:energySimulation}
}
\end{figure}

\Cref{fig:energySimulation}a shows the relation between the change in drop height and the corresponding change in dissipated energy by the simulated dampers for set \num{1}, \num{3} and \num{5} (continuous line for pure viscous, dashed for pure Coulomb damping). For the range of simulated drop heights, pure viscous and Coulomb dampers change the amount of dissipated energy with an almost linear dependence on the drop height. However, pure viscous damping has a slope closer to the \emph{full rejection} scenario (blue line in \Cref{fig:energySimulation}a), regardless of the set considered. In a step-down perturbation ($\Delta h > 0$ in \Cref{fig:energySimulation}a), pure viscous damping dissipates more of the additional energy $\Delta E_T$, while in a step-up perturbation ($\Delta h < 0$) it dissipates less energy than pure Coulomb damping. As such, the results show that a viscous damper can reject a step-down perturbation faster, e.g., within less hopping cycles, and it requires smaller correction by the active energy supply during a step-up perturbation. 

Adjusting the damping parameters allows to change the reaction to a perturbation (\Cref{fig:energySimulation}). Increasing the damping intensity, i.e., $d_v$ and $d_c$ from set \num{1} to \num{5}, allows to better match the \textit{full recovery} behaviour (blue line in \Cref{fig:energySimulation}a). However, this comes at the cost of a higher energy dissipation 
\mrkb{at the reference height, i.e., in absence of a ground perturbation (\Cref{tab:energy}, column `reference height'). Increasing the damping rate also affects the energetic advantage of viscous damping over Coulomb damping. \Cref{fig:energySimulation}b shows this in detail for a \emph{specific} step down perturbation $\left(\Delta h =\SI{2.5}{cm}\right)$: from set \num{1} to set \num{3}, the spread between the $\Delta E_D$ values of the viscous damper and the Coulomb damper increases (from \SI{8}{mJ} to \SI{18}{mJ}). However, the difference in dissipated energy $\Delta E_D$ slightly reduces from set \num{3} to set \num{5} (from \SI{18}{mJ} to \SI{16}{mJ}).}

\Cref{tab:energy} quantifies the previous findings by indicating the percentage of energy perturbation $\Delta E_T$ that each damping approach dissipates for $\Delta h = \pm \SI{2.5}{cm}$ and for all the tested sets of damping coefficients $d_v$ and $d_c$. The data further confirms the observations from  \Cref{fig:energySimulation}, showing that: 
\begin{enumerate}
\item \mrkb{within each set,} viscous damping outperforms Coulomb damping for all the simulated conditions - its dissipated energy is always the closest to \SI{100}{\%} of $\Delta E_T$, which means the closest to \emph{full rejection};
\item the energetic benefit of viscous damping over Coulomb damping, i.e., the spread in percentage values of $\Delta E_D/\Delta E_T$, 
\mrkb{\emph{does not monotonically increase}} with higher damping rates, i.e., moving from set 1 to 5. 
\end{enumerate}
Furthermore, \Cref{tab:energy} shows that for small damping rates, i.e., set \num{1}, viscous damping introduces only marginal benefits in energy management compared to Coulomb damping: $<10\%$ spread between the corresponding $\Delta E_D/\Delta E_T$ values.

\section{Hardware Description}\label{Hardware_description}
With the previous results from our numerical simulation in mind, we tested two technical implementations (\Cref{fig:damperImplementation}) to produce adjustable and viscous physical damping. We implemented a 2-segment leg hardware (\Cref{subfig:legDesign}) and mounted it to a vertical drop test bench to investigate the role of physical damping. The drop test bench produces velocity profiles during impact and stance phase similar to continuous hopping and allows us testing effective damping efficiently and repeatable.

\subsection{Rolling Diaphragm Damper}
\begin{figure*}[t]
\setcounter{subfigure}{0}
\centering
\subcaptionbox{Diaphragm damper\label{subfig:diaphragm}}
{\includegraphics[scale=1.]{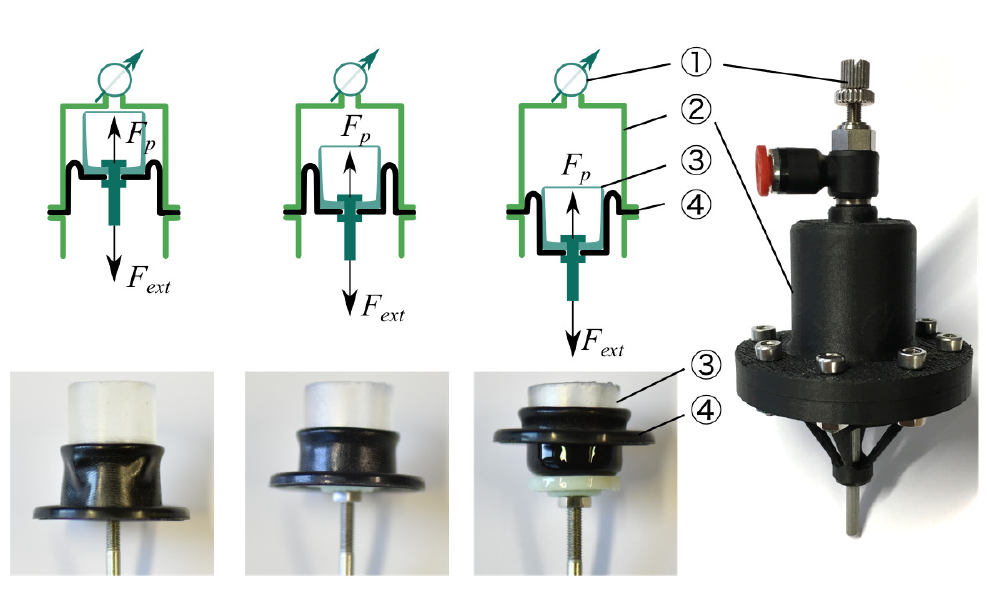}}
\hfill
\subcaptionbox{Hydraulic damper\label{subfig:hydraulic}}
{\includegraphics[scale=1.]{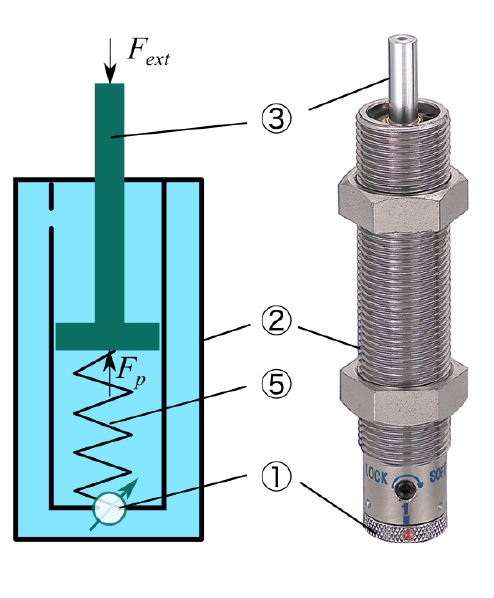}}
\caption{\textbf{(a)}-left-top: schematic of a diaphragm damper, illustrating the motion of rolling diaphragm, which includes an
adjustable orifice{\large \textcircled{\footnotesize 1}}, a
cylinder{\large \textcircled{\footnotesize 2}}, a
piston{\large \textcircled{\footnotesize 3}}, and a
rolling diaphragm{\large \textcircled{\footnotesize 4}}.
\textbf{(b)}-left: schematic of a hydraulic damper, fluid is sealed inside the  
cylinder{\large \textcircled{\footnotesize 2}} with an
recovery spring{\large \textcircled{\footnotesize 5}} to reset the
piston{\large \textcircled{\footnotesize 3}}.
}
\label{fig:damperImplementation}
\end{figure*}

The most common designs of viscous dampers are based on hydraulic or pneumatic cylinders (viscous damping) and can offer the possibility of regulating fluid flow by altering the orifice opening (adjustability). These physical dampers can display high Coulomb friction, caused by the mechanical design of the sliding seal mechanisms. Typically, the higher the cylinder pressure is, the higher the Coulomb friction exists. Ideally, we wanted to test one physical damper concept with the least possible amount of Coulomb friction. Inspired by the low-friction hydrostatic actuators \citep{whitney_low-friction_2014,whitney_hybrid_2016}, we designed a low-Coulomb damper based on a rolling diaphragm cylinder. Its cylinder is 3D printed from Onyx material. \Cref{subfig:diaphragm} illustrates the folding movement of this rolling diaphragm mounted on a piston. The rolling diaphragm is made of an elastomer shaped like a top hat that can fold at its rim. When the piston moves out, the diaphragm envelopes the piston. In the ideal implementation, only rolling contact between the diaphragm and the cylinder occurs, and no sliding contact. Hence, Coulomb friction between piston and cylinder is minimized. We measured $F_C\approx\SI{0.3}{N}$ of Coulomb friction for our rolling diaphragm cylinder, at low speed. 

Our numerical simulation results promoted viscous and adjustable damping for use in vertical leg-drop. By concept, both properties are satisfied by the diaphragm damper with an adjustable valve. When an external load $F_{ext}$ pulls the damper piston (\Cref{subfig:diaphragm}), the fluid inside the cylinder chamber flows through a small orifice, adjustable by diameter. This flow introduces a pressure drop  $\Delta P(t)$, whose magnitude depends on the orifice cross-section area $A_o$ and piston speed $v(t)$. As such, for a given cylinder cross section area $A_p$, the diaphragm damper reacts to an external load $F_{ext}$ by a viscous force $F_p(t)$ due to the pressure drop $\Delta P(t)$:
\begin{equation}
\label{eqnxx}
F_p(t) = A_p\, \Delta P(t) = A_p\, f(v(t),A_o) \end{equation}
We mounted a manually adjustable valve (SPSNN4, MISUMI) to set the orifice size $A_o$. For practical reasons (weight, leakage, complexity of a closed circuit with two cylinders) we used air in the diaphragm cylinder as the operating fluid, instead of liquid \citep{whitney_low-friction_2014,whitney_hybrid_2016}. Air is compressible, and with a fully closed valve the diaphragm cylinder also acts as an air spring. This additional functionality can potentially simplify the overall leg design. With the pneumatic, rolling diaphragm-based damper implementation, we focused on creating a light-weight, adjustable damper with minimal Coulomb friction, and air as operating fluid.

\vspace{1.5cm}
\subsection{Hydraulic Damper} \label{meth:HydrDamper}
In the second technical implementation we applied an off-the-shelf hydraulic damper (1214H 
or 1210M, MISUMI,  \Cref{subfig:hydraulic}), i.e., a commercially available solution for adjustable and viscous damping. Tested against other hydraulic commercial dampers, we found these specific models to have the most extensive range of adjustable viscous damping and the smallest Coulomb friction ($F_C\approx\SI{0.7}{N}$). Similarly to the diaphragm damper, these hydraulic dampers produce viscous damping by the pressure drop at the adjustable orifice. The operating fluid is oil, which is in-compressible. Hence, the hydraulic damper should not exhibit compliant behavior. Other than the diaphragm damper, the hydraulic damper produces damping force when its piston is pushed, not pulled. This design also includes an internal spring to recover the piston position when unloaded. In sum, the hydraulic damper features high viscous damping, no air-spring effect, and a higher Coulomb friction compared the custom-designed pneumatic diaphragm damper.
\\

\subsection{Articulated Leg Design}\label{sec:articulatedLeg}
The characteristics of a viscous damper strongly depend on the speed- and force-loading profile imposed at its piston, because of the complex interaction of fluid pressure and compression, viscous friction, and cavitation \citep{Dixon2008}. We implemented a hardware leg to test our two physical dampers at loading profiles (speed, force) similar to legged hopping and running.

The 2-segment hardware leg (\Cref{subfig:legDesign}) is designed with a constant spring and damper lever arm, parameters are provided in \Cref{tab:experimentPara}. In all experiments with the 2-segmented leg, the leg spring provides elastic joint reaction forces. Dampers are swapped in and out in a modular fashion, depending on the experimental settings. The 2-segment leg design parameters are identical to those in our simulation model (\Cref{tab:designPara}). A compression spring{\large \textcircled{\footnotesize 8}} is mounted on the upper leg segment{\large \textcircled{\footnotesize 13}}. When the leg flexes, the spring is charged by a spring cap{\large \textcircled{\footnotesize 7}} coupled to a cable{\large \textcircled{\footnotesize 10}} attached to the lower leg. Either damper{\large \textcircled{\footnotesize 15}}{\large \textcircled{\footnotesize 16}} is fixed on a support{\large \textcircled{\footnotesize 6}} on the upper segment{\large \textcircled{\footnotesize 13}}. The support{\large \textcircled{\footnotesize 6}} can be moved within the upper segment{\large \textcircled{\footnotesize 13}}, to adjust the cable{\large \textcircled{\footnotesize 9}} pretension. Cables{\large \textcircled{\footnotesize 9}}{\large \textcircled{\footnotesize 10}} link the damper piston{\large \textcircled{\footnotesize 3}} and the spring{\large \textcircled{\footnotesize 8}} to the knee pulley{\large \textcircled{\footnotesize 11}}, which is part of the lower segment{\large \textcircled{\footnotesize 12}}.

During the leg flexion, the cable under tension transmits forces instantly to the spring and damper. Spring and damper forces counteract the knee flexion. During leg extension, the spring releases energy, while the damper is decoupled due to slackness of the cable. We included a hard stop into the knee joint to limit the maximum leg extension, and achieve a fixed leg length at impact. At maximum leg flexion at high leg loading, segments can potentially collide. We ensured not to hit either hard stops during the drop experiments. The hydraulic damper{\large \textcircled{\footnotesize 16}} requires a reverse mechanism{\large \textcircled{\footnotesize 14}}, since its piston requires compression to work. The piston of the diaphragm damper{\large \textcircled{\footnotesize 15}} was directly connected to the knee pulley. The diaphragm damper{\large \textcircled{\footnotesize 15}} included no recovery spring{\large \textcircled{\footnotesize 5}}, hence we reset the piston position manually after each drop test. In sum, different spring-damper combinations can be tested with the 2-segment leg setup.
\mrkb{Note that the here shown hardware leg has no actuation. If a motor would actuate the knee joint, in parallel mounted to the spring and the damper, the damper would share the external impact load, and consequently reduce an impact at the motor.}

\vspace{1cm}
\subsection{Experimental set-up, data sampling and processing} 
\label{sec:experimentalsetup}
We implemented an experimental setup for repetitive measurements (\Cref{subfig:legPhoto}). A drop bench was used to constrain the leg motion to a single vertical degree of freedom, and linear motion. This allowed us to fully instrument the setup (slider position, and vertical ground reaction forces, GRF), and ensured repeatable conditions over trials. Adjusting the drop height allowed us setting the touch-down speed. A linear rail (SVR-28, MISUMI) was fixed vertically on a frame. The upper leg segment was hinged to a rail slider. The rail slider was loaded with additional, external weights, simulating different robot masses. We set the initial hip angle $\alpha_0$ to align the hip and foot vertically. A hard stop ensured that the upper leg kept a minimum angle $\alpha > \alpha_0$. 

\mrkb{Two} sensors measured the leg dynamics: the body position $y$ and the vertical ground reaction force are recorded by a linear encoder (AS5311, AMS) and a force sensor (K3D60a, ME, amplified with \num{9326}, Burster), respectively (\Cref{subfig:legPhoto}). The duration from touch-down to mid-stance is very short, typically $t \leq \SI{100}{ms}$, and high-frequency data sampling was required. The encoder data was sampled by Raspberry Pi 3B+ with $\textrm{f}=\SI{8}{kHz}$ sampling rate. Force data were recorded by an Arduino Uno, with a 10-bit internal ADC at \SI{1}{kHz} sampling rate. A high-speed camera (Miro Lab \num{110}, Phantom) recorded the drop sequence at $\textrm{f}=\SI{1}{kHz}$ sampling rate. We performed ten trials for each test condition.  Sensor data was processed with MATLAB (the MathWorks, Natick, MA). Data was smoothed with a moving average filter, with a filter span of \num{35} samples for encoder data, and \num{200} samples for force data. Repeated experiments of the same test condition are summarized as an envelop defined by the average $\pm$ the standard deviation of the filtered signals. 

\section{Hardware experiments and results}\label{Experiments and results}
\begin{figure*}[ht]
\centering
\includegraphics[width = 1.0\linewidth]{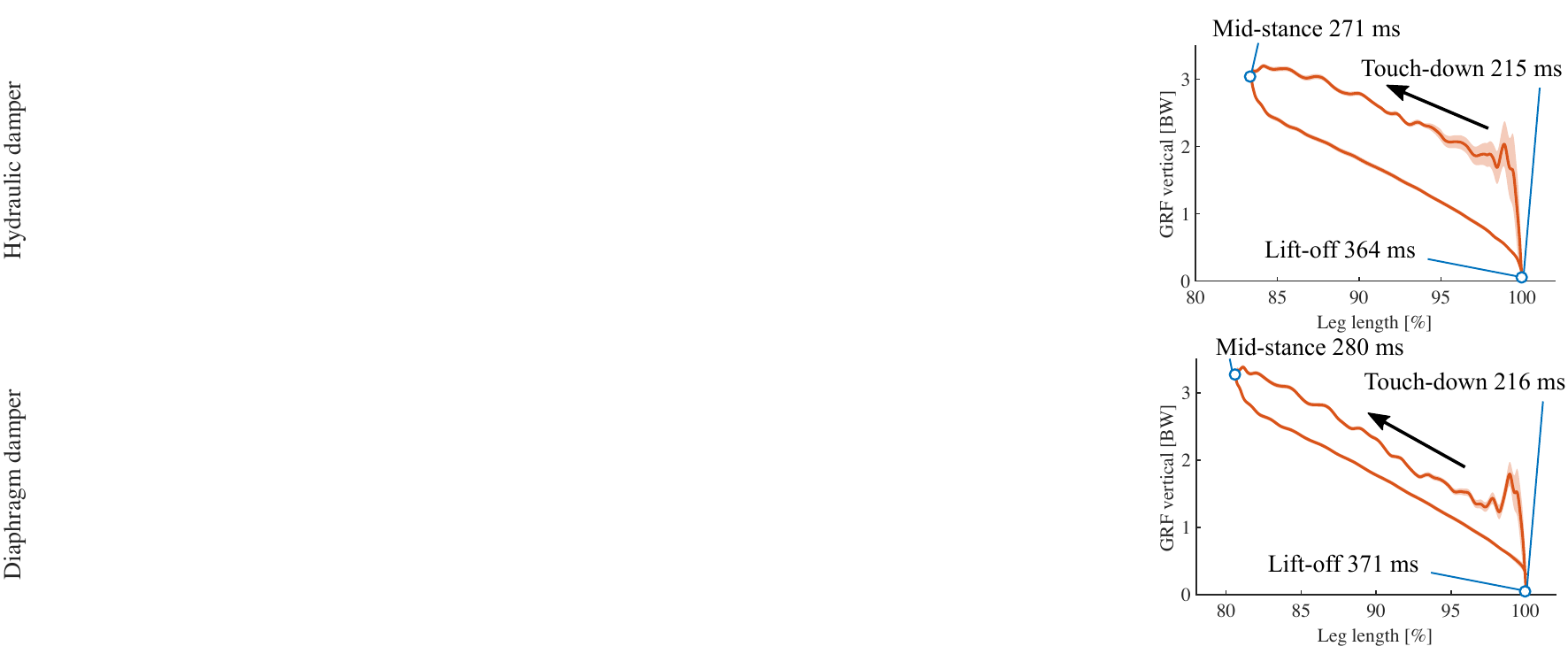}
\caption{High-speed snapshots of drop experiments starting from release to second touchdown. Leg with hydraulic damper is shown on the top row, leg with diaphragm damper the bottom row. Depicted are from left to right: release, touchdown, mid-stance, lift-off, apex, second touchdown. The right plots illustrate the timing of the events corresponding to the snapshots.
}
\label{fig:highSpeedSnapshots}
\end{figure*}

In the drop experiments, we characterize both the hydraulic and diaphragm dampers, and the 2-segment springy leg \mrkb{(\Cref{fig:highSpeedSnapshots})}.
\mrkb{We chose three orifice settings (labeled as a, b, and c) for each damper, and focus on the effects of viscous damping and adjustable dissipation of energy in the hardware setup. \Cref{tab:experimentPara} lists an overview of the drop tests, and its settings (drop height, weight, orifice setting, damper type).}
\mrkb{To emphasize the fundamental differences between the damper designs, we compare only one model of the hydraulic damper (1214H) to the diaphragm damper (\Cref{subsec:Results_IsolatedDamper} - \Cref{subsec:Results_Adjustability}), and show the potential of the second hydraulic damper (1210M) in \Cref{subsec:Results_Damper_Selection}.} Videos of the experiments can be found in the supplementary material, and online~\footnote{\url{https://youtu.be/F00Sma2BQ4c}}.

\begin{table*}[ht]
\fontsize{9}{11}\selectfont
\caption{Drop test settings for experiments}
\label{tab:experimentPara}
\begin{center}
\begin{tabular}{llccc}
\multicolumn{2}{c}{\textit{Drop test setup}} & \textit{Drop height} & \textit{Drop weight} & \textit{Orifice} \\
& Fig. &  [\si{cm}]       &   [\si{g}]      &  [$\sim$] \\
\hline
\multirow{3}{*}{Damper (1214H)}   & \ref{fig:Fv_height}  & \textbf{3, 5, 7}           & 280                   & b\\
     & \ref{fig:Fv_valve}             & 5             & 280         & \textbf{a, b, c}\\
         & \ref{fig:Fv_weight}              & 3             & \textbf{280, 620}      & b\\
\hline
\multirow{2}{*}{Damper (1214H, diaphragm) \& leg}     & \ref{fig:Fl_ind}, \ref{fig:Fl_diaph}        & 14    & 408      & c\\
         & \ref{fig:Fl_spring}           & 14    & 408      & damper detached\\
\hline
\multirow{2}{*}{Damper \& leg (simulation)}    & \ref{fig:tunability_ind}, \ref{fig:tunability_diaph}    &    14   &  408   &  \textbf{a, c}\\
         & \ref{fig:tunability_sim}       &    14   &  408   &  \textbf{viscous, Coulomb}\\
\hline
		{Damper (1210M) \& leg}     & \ref{fig:1210M}        & 14    & 408      & \textbf{a, b}
\end{tabular}
\end{center}
\end{table*} 

\subsection{Isolated damper drops, evaluation}\label{subsec:Results_IsolatedDamper}
In this experiment we characterized the hydraulic damper by dropping it under changing conditions of the instrumented drop setup, without mounting it to the 2-segment leg. The experimental setup allows differentiating effects, compared to the 2-segment leg setup, and to emphasize the viscous damper behavior of the off-the-shelf component. We also applied the results to estimate the range of damping rates available with changing orifice settings. The hydraulic damper was directly fixed to the rail slider into the drop bench (\Cref{sec:experimentalsetup}). The piston pointed downwards. We measure the vertical ground reaction force to determine the piston force, and we recorded the vertical position of the slider over time, to estimate the piston speed after it touches the force sensor. 

\begin{figure*}[ht]
\setcounter{subfigure}{0}
\centering
\subcaptionbox{Three drop heights\label{fig:Fv_height}}
{\includegraphics[scale=1.0]{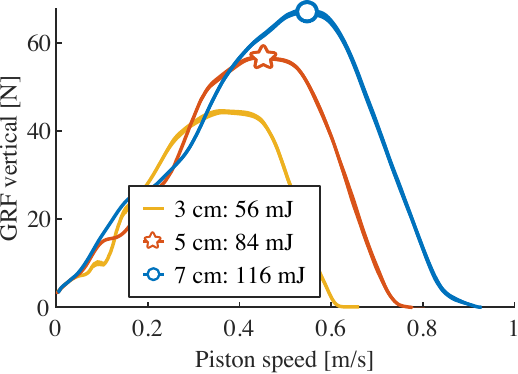}
\llap{\raisebox{2cm}
{\includegraphics[height=2cm]{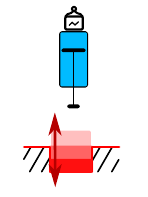}
}}}\hfill
\subcaptionbox{Three valve settings\label{fig:Fv_valve}}
{\includegraphics[scale=1.0]{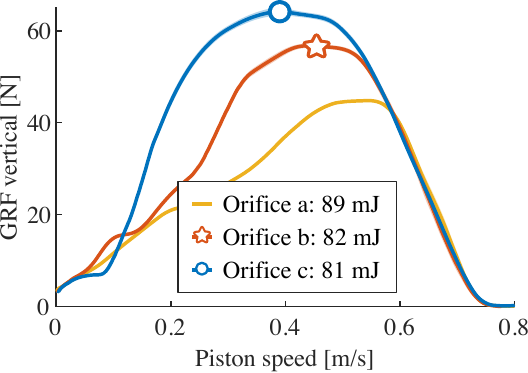}
\llap{\raisebox{2cm}
{\includegraphics[height=2cm]{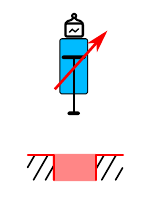}
}}
}\hfill
\subcaptionbox{Two system weights\label{fig:Fv_weight}}
{\includegraphics[scale=1.0]{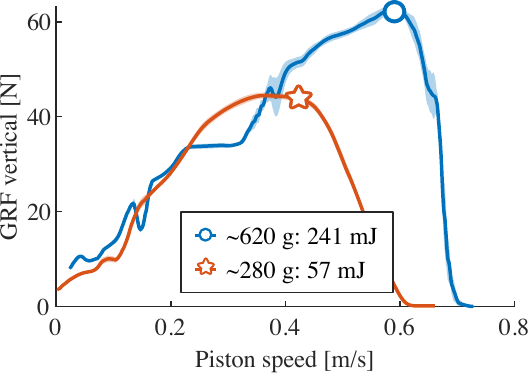}
\llap{\raisebox{2cm}
{\includegraphics[height=2cm]{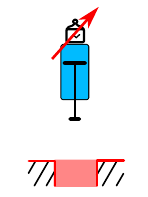}
}}}
\caption{\textbf{Characterizing the hydraulic damper} A single damper (not leg-mounted) drops onto the force sensor. \num{10} repeated experiments are plotted as an envelop, defined by the average \SI{\pm 95}{\%} of the standard deviation data. The curves are read from right to left, i.e.\,from touch-down at maximum speed to zero speed at rest, also corresponding to the maximum damper compression. \textbf{(a)} \SI{280}{g} drop mass with medium orifice in \num{3} drop heights. \textbf{(b)} \SI{280}{g} drop mass with \SI{5}{cm} drop height in \num{3} orifice settings. 
\textbf{(c)} \SI{3}{cm} drop height with medium orifice in \num{2} drop weights.}
\label{fig:damper_eval}
\end{figure*}

\Cref{fig:damper_eval} shows the force-speed profiles for drop tests with different drop heights (\Cref{fig:Fv_height}), orifice settings (\Cref{fig:Fv_valve}), and drop loads (\Cref{fig:Fv_weight}). Data lines in \Cref{fig:damper_eval} should be interpreted from high speed (impact, right side of each plot) to low speed (end of settling phase, \SI{0}{m/s}, left). The time from impact to peak force (right slope of each plot) is ($\approx$\SI{24}{ms}), while the negative work (shown in legends) was mainly dissipated along the falling slope in the much longer-lasting settling phase after the peak (left slope of each plot, $\approx$\SI{200}{ms}).

The results from tests with drop heights from \SI{3}{cm} to \SI{7}{cm} show viscous damping behavior in the settling phase after peak force (left slope), with higher reaction forces at higher piston speeds with higher dissipation, ranging from \SI{45}{N} for maximum speeds of \SI{0.6}{m/s} with \SI{56}{mJ} to \SI{65}{N} at \SI{0.9}{m/s} with  
\mrkb{\SI{116}{mJ}}. The piston force almost linearly depends on the piston speed (\Cref{fig:Fv_height}).

Changing the orifice setting at a constant drop height resulted in different settling slopes (\Cref{fig:Fv_valve}).
Applying a least-squares fit on the left-falling settling slope, we estimate an adjustable damping rate between \SI{91}{Ns/m} \mrkb{and} \SI{192}{Ns/m}. The dissipated energy changes from \SI{89}{mJ} to \SI{81}{mJ}, respectively. Hence adjusting the orifice setting has an effect on the damping rate and the dissipated energy in the isolated hydraulic damper, but not as we intuitively expected.  

\mrka{We interpret the rising slope in the impact phase (right part of each curve, \Cref{fig:Fv_height,fig:Fv_valve}) as a build-up phase; the hydraulic damper takes time ($\approx$\SI{24}{ms}) to build up its internal viscous flow and the related piston movement, after the piston impact. With} heavier weights (\SI{620}{g} = heavy, \SI{280}{g} = light, \Cref{fig:Fv_weight}), the impact phase equally lasts $\approx$\SI{24}{ms}. After the impact phase with heavy weight, the damper shows the same damping rate in the settling phase, in form of an equal left slope.

\mrkb{Similar drop tests for the evaluation of the isolated diaphragm damper were not possible since the orientation of the internal diaphragm only permits to pull the piston. In the following section, we test the diaphragm (connected by a piston reverse mechanism) and the hydraulic damper directly on the 2-segment leg structure.}

\vspace{1cm}
\subsection{Composition of dissipated energy}\label{subsec:Results_Composition_energy_diss}
We performed drop tests of two damper configurations: one off-the-shelf hydraulic damper, and custom-made pneumatic damper, each mounted in parallel to a spring at the 2-segment leg (\Cref{sec:articulatedLeg}, \Cref{subfig:legDesign}), to quantify the effect of viscous damping for drop dynamics similar to legged hopping.

For each drop, the effective dissipated energy $E_\mathrm{effective}$ was computed by calculating the area enclosed by the vertical GRF-leg length curve from touch-down to lift-off \citep{Josephson_1985}, i.e., the work-loop area.
These work-loops are to be read counter-clockwise, with the rising part being the loading during leg flexion, and the falling part being the unloading, due to spring recoil.
$E_\mathrm{effective}$ does not only consist of the viscous loss $E_\mathrm{viscous}$ due to the damper, but also Coulomb friction loss in the leg ($E_\mathrm{cfriction}$) and the impact loss $E_\mathrm{impact}$ due to unsprung masses:
\begin{equation}
\label{eqn10}
E_\mathrm{effective} = E_\mathrm{cfriction} + E_\mathrm{impact} + E_\mathrm{viscous}. \end{equation}

We propose a method to indirectly calculate the contribution of viscous damping, by measuring and eliminating effects from Coulomb friction, and unsprung masses.

To quantify the Coulomb friction loss $E_\mathrm{cfriction}$, we conducted `slow drop' tests. The mechanical setup is identical to `free drops' test, where the leg is freely dropped from a fixed height. However, in the `slow drop' experiment the 2-segment leg is lowered manually onto the force plat, contacting and pressing the leg-damper-spring system onto the force plate. At slow speed only Coulomb friction in joints and damper act, but no viscous damping or impact losses occur. Consequently the dissipated energy calculated from the size of the work loop is due to Coulomb friction losses $E_\mathrm{cfriction}$.

\begin{figure*}[ht]
\setcounter{subfigure}{0}
\centering
\subcaptionbox{Hydraulic damper and spring\label{fig:Fl_ind}}
{\includegraphics[scale=1.0]{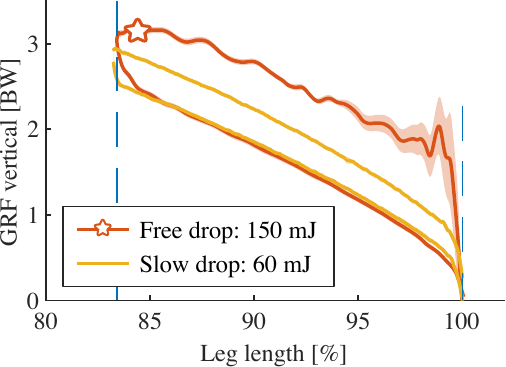}
    \llap{\raisebox{2.5cm}
    {\includegraphics[height=1.2cm]{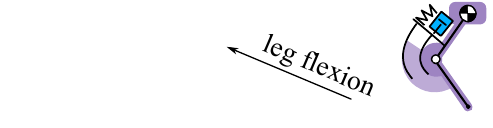}
    }}
}\hfill
\subcaptionbox{Diaphragm damper and spring\label{fig:Fl_diaph}}
{\includegraphics[scale=1.0]{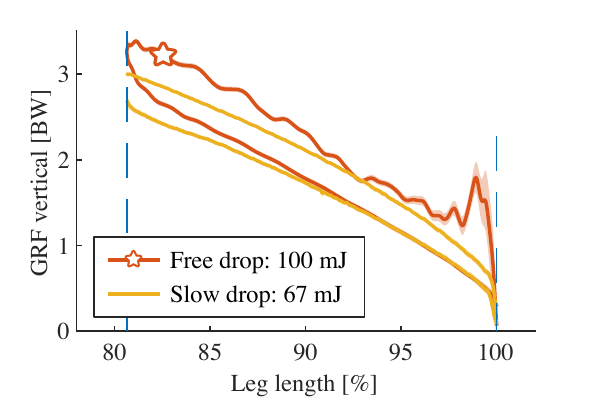}
    \llap{\raisebox{2.5cm}
    {\includegraphics[height=1.2cm]{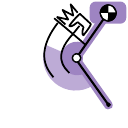}
    }}
}\hfill
\subcaptionbox{Spring only\label{fig:Fl_spring}}
{\includegraphics[scale=1.0]{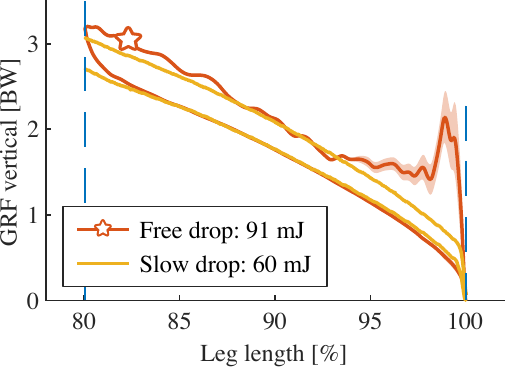}
    \llap{\raisebox{2.5cm}
    {\includegraphics[height=1.2cm]{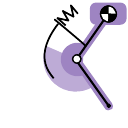}
    }}}
\caption{\textbf{Characterizing the contribution of velocity-dependent damping:} Vertical GRF versus leg length change, a 2-DOF leg with damper/spring drops onto the force sensor: Three different hardware configurations were tested, for slow and free drop speeds on the vertical slider. Yellow data lines indicate slow-motion experiments. Experiments `start' bottom right, at normalized leg length \SI{100}{\%}. Reading goes counter-clockwise, i.e.\,from touch-down to mid-stance is indicated by the upper part of the hysteresis curve, while the lower part indicates elastic spring-rebound, without damper contribution.}
\label{fig:Fl}
\end{figure*}

To identify the impact loss $E_\mathrm{impact}$, we remove the viscous component first by detaching the damper cable on the setup. A `free drop' test in this spring only condition measures the contribution of friction loss $E_\mathrm{cfriction}$ and impact loss $E_\mathrm{impact}$ combined. A `slow drop' test of the same setup is able to quantify the friction loss $E_\mathrm{cfriction}$. The impact loss $E_\mathrm{impact}$ is therefore estimated as the energy difference between `free drop' and `slow drop' in the spring-only condition (\Cref{fig:Fl_spring}). Since the effective dissipated energy $E_\mathrm{effective}$ is directly measured, and the friction loss $E_\mathrm{cfriction}$ and impact loss $E_\mathrm{impact}$ are obtained separately, the viscous loss $E_\mathrm{viscous}$ can be computed according to \mrkb{\Cref{eqn10}}.

\Cref{fig:Fl_ind,fig:Fl_diaph} show the `free drop' and `slow drop' results of the hydraulic damper and diaphragm damper, respectively. Both drop heights are \SI{14}{cm}, at identical orifice setting. We calculated the negative work of each work-loop (range indicated by the two vertical dash lines), as shown in \Cref{fig:Fl}. To provide an objective analysis, the work-loop area of each `slow drop' (manual movement) was cut to the maximum leg compression of the corresponding `free drop' condition. The dissipated energy of the leg-mounted hydraulic damper is \SI{150}{mJ} and \SI{60}{mJ} for `free drop' and `slow drop', respectively, and \SI{100}{mJ} and \SI{67}{mJ} for the diaphragm damper, respectively. According to \Cref{fig:Fl_spring}, the impact loss $E_\mathrm{impact}$ due to unsprung masses play a large role, accounting for \SI{31}{mJ}. The viscous loss $E_\mathrm{viscous}$ of the hydraulic and the diaphragm damper are \SI{59}{mJ} and \SI{2}{mJ}, respectively.

\subsection{Adjustability of dissipated energy}\label{subsec:Results_Adjustability}
\begin{figure*}[htpb]
\setcounter{subfigure}{0}
\centering
\subcaptionbox{Hydraulic damper and spring\label{fig:tunability_ind}}
{\includegraphics[scale=1.0]{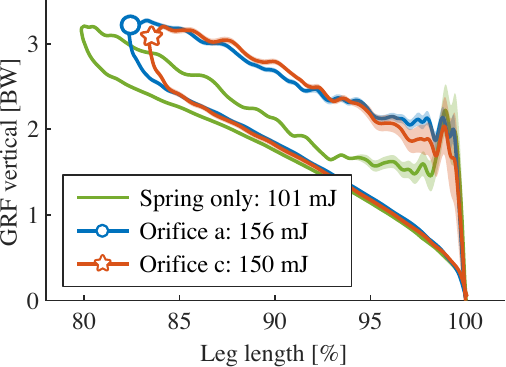}
    \llap{\raisebox{2.5cm}
    {\includegraphics[height=1.2cm]{thumbnails_leg_springDamper_hydraulic}
    }}
}\hfill
\subcaptionbox{Diaphragm damper and spring\label{fig:tunability_diaph}}
{\includegraphics[scale=1.0]{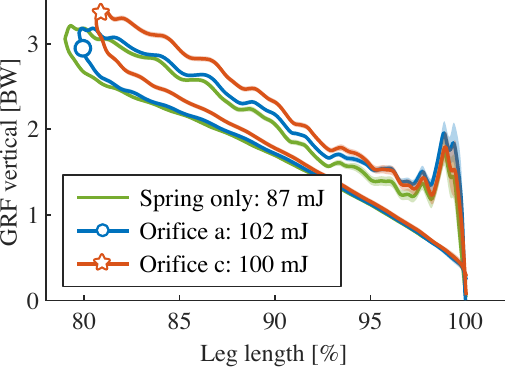}
    \llap{\raisebox{2.5cm}
    {\includegraphics[height=1.2cm]{thumbnails_leg_springDamper_diaphragm}
    }}
}\hfill
\subcaptionbox{Simulation\label{fig:tunability_sim}}
{\includegraphics[scale=1.0]{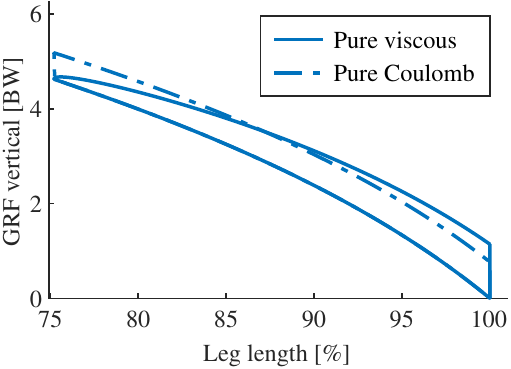}}
\caption{\textbf{Adjustability and tunability of damping:} Vertical GRF vs leg length change, a 2-segment leg with damper and spring drops onto the force sensor. Two different hardware configurations were tested, for different orifice settings. \textbf{(a) and (b)} show the result from hydraulic damper and diaphragm damper respectively, where the \mrkb{green} data lines indicate the leg drop without damper for comparison. \textbf{(c)}: \emph{Simulated} approximation of hydraulic damper orifice a by a pure viscous and a Coulomb damper. Damping coefficients are chosen to allows same dissipated energy, i.e., $E_{D_0} = \SI{156}{mJ}$: respectively --- pure viscous damper: $d_c$ = \SI{0}{N} and $d_v$ = \SI{51}{Ns/m}; pure Coulomb damper: $d_c$ = \SI{13.2}{N} and $d_v$ = \SI{0}{Ns/m}. None of the two curves can fully capture the work-loop of hydraulic damper.}
\label{fig:tunability}
\end{figure*}

We tested the adjustability of energy dissipation during leg drops by the altering orifice setting for each leg-mounted damper, and quantified by calculating the size of the resulting work-loops. The drop height was fixed to \SI{14}{cm} and we used \num{2} orifice settings. The identical same set-up but in spring-only configuration (damper cables detached) was tested for reference. Work-loop and corresponding effective dissipated energies are illustrated in \Cref{fig:tunability_ind,fig:tunability_diaph}. The hydraulic damper-mounted leg dissipated \SI{156}{mJ} and \SI{150}{mJ} energy on its two orifice settings, the pneumatic diaphragm damper dissipated \SI{102}{mJ} and \SI{100}{mJ}. In \Cref{fig:tunability_sim}, we display results from the numerical model introduced in \Cref{Sec:simulation} to estimate the work-loop shape that either a pure viscous or pure Coulomb damper would produce, if dissipating the same amount of energy as the hydraulic damper with orifice-a (\Cref{fig:tunability_ind}). We set the damping coefficients of our numerical model to $E_{D_0} \approx\SI{156}{mJ}$, so that: $\left(d_v, d_c\right) = \left(\SI{51}{Ns/m},\SI{0}{N}\right)$ for pure viscous damping; and $\left(d_v, d_c\right) = \left(\SI{0}{Ns/m},\SI{13.2}{N}\right)$ for pure Coulomb damping. Work-loops from the numerical simulation differ notably from the experimental data, suggesting that neither the hydraulic or diaphragm damper can easily be approximated as pure viscous or pure Coulomb  dampers. Both work loops in \Cref{fig:tunability_sim} present about equal amount of dissipated energy. Yet, both differ greatly due to their underlying damping dynamics, visible in their unique work-loop shapes.
Their individual characteristics are different enough to uniquely identify pure viscous or pure Coulomb dampers, from numerical simulation.

\vspace{1cm}
\subsection{Damper selection choices} \label{subsec:Results_Damper_Selection}
\begin{figure}[ht]
\centering
\includegraphics[scale = 1.0]{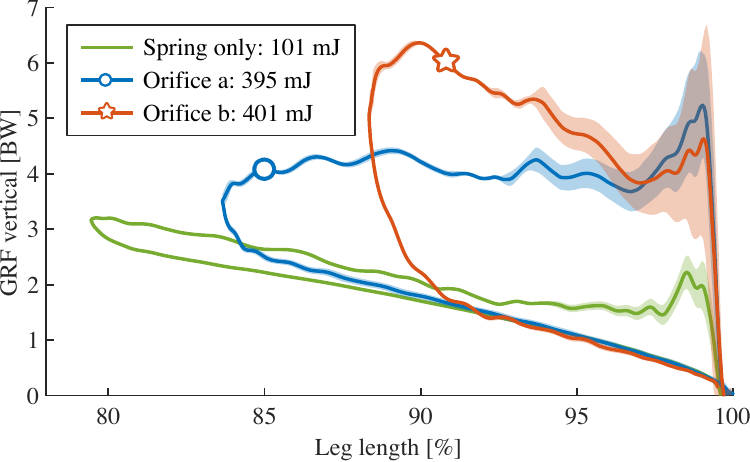}
\caption{\textbf{Higher energy dissipation with a different model of the hydraulic damper (1210M):} Vertical GRF vs.\,leg length change, a 2-DOF leg with a parallel damper and spring drops onto the force sensor. Two damper orifice settings were tested (blue, red curves). The two resulting curves are compared with the spring-only configuration, provided as reference.}
\label{fig:1210M}
\end{figure}

\mrkb{In accordance with the simulation results, we aim to use a viscous damper to dissipate energy introduced by a ground disturbance. How much energy could be dissipated by the damper, depended mainly on the selected viscous damper, and only to a limited degree on the orifice setting. Results from the hydraulic damper 1214H showed significant energy dissipation capabilities: $\approx$\SI{11}{\%} of the system's total energy (\SI{59}{mJ} of \SI{560}{mJ}) were dissipated (\Cref{fig:tunability_ind} at orifice setting 'c' and \Cref{tab:energyComposition} ). At the drop, in sum \SI{150}{mJ} (\SI{27}{\%}) of the leg's system energy were lost, due to Coulomb friction in the joints, impact dynamics, and viscous damping losses. Other dissipation dynamics are feasible, by selecting appropriate dampers. We tested a second hydraulic damper (1210M, MISUMI) under equal conditions and compared it to damper-1214H. The two applied orifice settings changed the observed work loop largely by shape, and little by area (\Cref{fig:1210M}). The damper-1210M dissipated $\approx$\SI{60}{\%} system energy, and the leg lost in sum (viscous+Coulomb+impact) \SI{72}{\%} of its system's energy during that single drop. At other orifice settings, we observed over-damping; the 1210M-spring leg came to an early and complete stop, and without rebound (data not shown here due to incomplete work loop).}

\begin{figure*}[htpb]
\setcounter{subfigure}{0}
\centering
\subcaptionbox{Vertical GRF over time\label{fig:time_GRF}}
{\includegraphics[scale=1.0]{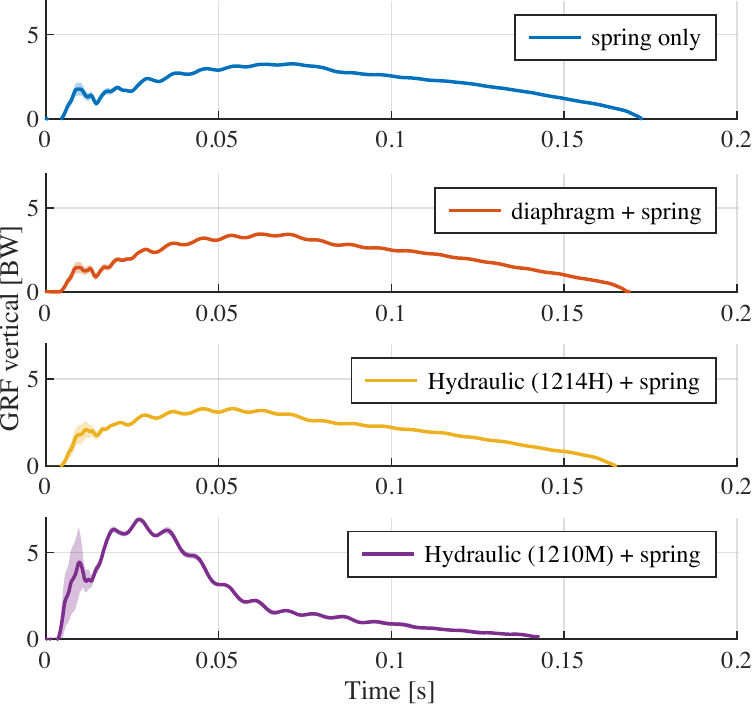}}
\hfill
\subcaptionbox{Vertical Impulse over time\label{fig:time_Impulse}}
{\includegraphics[scale=1.0]{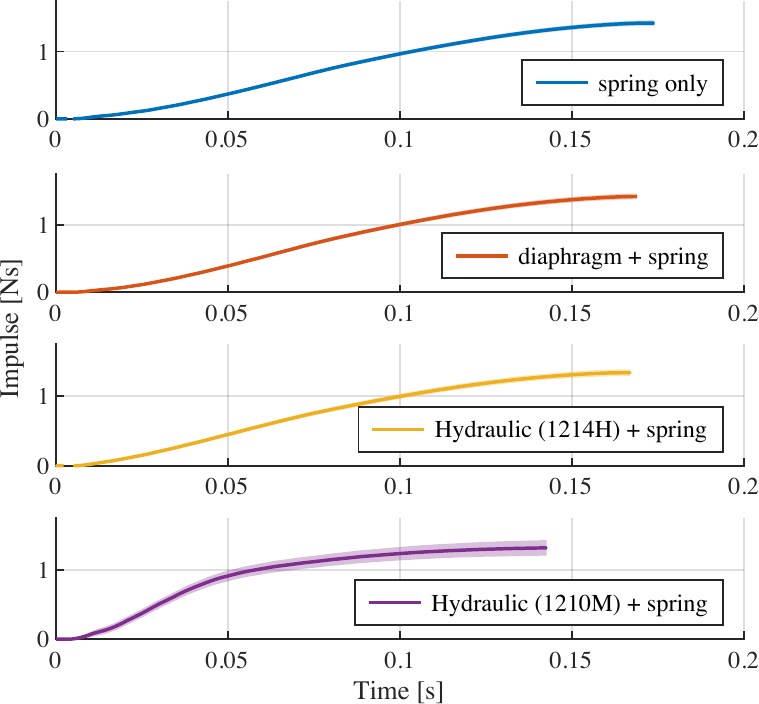}}
\caption{\mrka{\textbf{Ground reaction forces and the corresponding, instantaneous impulse for leg drop experiments}.
The corresponding work curves are provided in \Cref{fig:Fl,fig:tunability,fig:1210M}.
}}
\label{fig:time-plot}
\end{figure*}

\mrkb{For comparison, time plots of the vertical GRF and the impulse at stance phrase are shown in \Cref{fig:time-plot}. The energy composition (Equation \ref{eqn10}) is provided in \Cref{tab:energyComposition}. The 'spring only' data correspond the curves in \Cref{fig:Fl_spring}. The diaphragm+spring data correspond to `orifice c' in \Cref{fig:tunability_diaph}. The hydraulic (1214H)+spring data correspond to `orifice c' in \Cref{fig:tunability_ind}. The hydraulic (1210M)+spring data correspond to `orifice b' in \Cref{fig:1210M}. Among the tested dampers, the hydraulic 1210M damper showed the largest vertical GRF; peak vertical GRF of \SI{6.3}{BW} are observed, almost twice as much as the `spring only' case. The viscous dampers 1214H and 1210M shifted the peak of their legs' vertical GRF to an earlier point in time, compared to the spring-leg and the spring+diaphragm-leg (\Cref{fig:time-plot}).} 

\begin{table*}[ht]
\small
\caption{%
\mrkb{\textbf{Leg drop experiments and their individual energetic losses per drop.}
The system's initial potential energy is \SI{560}{mJ}.
$E_\mathrm{effective}$: sum of all energetic losses visible as the area of the hysteresis curve, i.e. in \Cref{fig:Fl},
$E_\mathrm{cfriction}$: negative work dissipated by Coulomb friction,
$E_\mathrm{impact}$: energetic losses from impact (unsprung mass).
The negative work dissipated by viscous damping in the physical damper is $E_\mathrm{viscous}$.
The corresponding work curves are provided in \Cref{fig:Fl,fig:tunability,fig:1210M}.
}}
\label{tab:energyComposition}
\begin{center}
\begin{tabular}{l|rrrr}
Drop test setup & $E_\mathrm{effective}$ & $E_\mathrm{cfriction}$ & $E_\mathrm{impact}$ & $E_\mathrm{viscous}$ \\
 & [\si{mJ}] & [\si{mJ}] & [\si{mJ}] & [\si{mJ}] \\
\hline
Spring only              	& 91  	& 60    & 31   	& 0\\
Diaphragm + spring 			& 100 	& 67	& 31    & 2\\
Hydraulic 1214H + spring    & 150	& 60	& 31  	& 59\\
Hydraulic 1210M + spring    & 401	& 38	& 31 	& 332\\
\end{tabular}
\end{center}
\end{table*} 

\section{Discussion}
A primary objective of this study was to test how physical dampers could be exploited for locomotion tasks by characterizing multiple available technical solutions. Our numerical model predicted three crucial aspects: (1) a pure viscous damper generally performs better than a pure Coulomb damper (\Cref{fig:energySimulation}); (2) higher damping rates result in better rejection of ground disturbances (\Cref{fig:energySimulation}a), however at the cost of higher dissipation at reference height (\Cref{tab:energy}); (3) characteristic work loop shapes for pure viscous and Coulomb damper during leg-drop (\Cref{fig:tunability_sim}). Our hardware findings show that neither of the tested physical dampers approximates as pure viscous or pure Coulomb dampers. The experiments also suggest that the mapping between dissipated energy and damping rates is concealed by the dynamics of the impact and the non-linearity of the force-velocity characteristics of the leg in the stance phase. Therefore, it is vital to test damping in a real leg at impact because the behavior is not merely as expected from the data sheets and the simple model. 

\Cref{fig:damper_eval} characterizes how the hydraulic damper dissipates energy during a free drop. The experimental results show that the dissipated energy of the hydraulic damper scales with drop height (\Cref{fig:Fv_height}) and weight (\Cref{fig:Fv_weight}), but less intuitively, it reduces with increasing damping rates (\Cref{fig:Fv_valve}).
This can be partially interpreted in the context of an ideal viscous damper (as in \Cref{eq:damper_torque_def}), but linear) for which the effective dissipated energy $E_\mathrm{effective}$ would be calculated as in,
\begin{equation}
    E_\mathrm{effective} = \int F_{p}(t)dy_p = \int \left(d_v \cdot v_p(t)\right) dy_p
    \label{eqn:damperEnergy_formula}
\end{equation}
where $F_{p}(t)$ is the damper piston force and $y_p$ is the piston displacement, $v_p(t)$ the corresponding velocity. When increasing the drop height, the velocity at impact is increased, so is $v_p(t)$. With the assumption of \Cref{eqn:damperEnergy_formula}, this results in higher damping forces $F_{p}(t)$, and thus, dissipated energy $E_\mathrm{effective}$, as seen in \Cref{fig:Fv_height}. The heavier drop weight leads to slower deceleration. Therefore the velocity profile $v_p(t)$ is increased, which also leads to higher dissipation $E_\mathrm{effective}$ (\Cref{fig:Fv_valve}). An orifice setting of high damping rate will increase the damping coefficient $d_v$. However, the velocity profile $v_p(t)$ is expected to reduce due to higher resistance. This simple analogy shows that the coupling between damping coefficient $d_v$ and velocity profile $v_p(t)$ makes it difficult to predict the energy dissipation by setting the orifice and serves as an interpretation of why adjusting the orifice generates a relatively small adjustment of \SI{10}{\%} (\SI{81}{mJ}-\SI{89}{mJ}) of the dissipated energy. Also, the impact phase (time for the damper to output its designed damping force under sudden load) introduces additional non-linearity to the output force profile. Overall, the results in \Cref{fig:damper_eval} indicate that we can approximate the damping force produced by the hydraulic damper to be viscous and adjustable--- as such dampers are typically designed \citep{Dixon2008}---, but the mapping of energy dissipation to orifice setting is difficult to predict in a dynamic scenario. 

\mrka{The approximation as a linear, velocity dependent damper allows us to rapidly estimate energy dissipation in simulation, over a range of parameters. However, the exact mapping of the hardware leg/spring/damper energy dissipation to orifice setting is difficult to predict, when basing the estimation only on the isolated-damper drop experiments from \Cref{fig:damper_eval}. Instead, the leg/spring/damper experiments show that the energetic losses from the impact remove \SI{31}{mJ} energy, compared to \SI{59}{mJ} damper losses. The high amount of force oscillations at impact (up to \SI{~1}{BW}, Fig. 8a) during the first \SI{3}{\%} leg length change leads us to believe that these impact oscillations move the damper’s dynamic working range, i.e., its resulting instantaneous force and velocity. The oscillations are likely caused by unsprung mass effects of the leg/spring/damper structure, and could not be captured in an isolated-damper setup, or---at least not easily---in a simulation.}

The work loops of leg drop experiments (\Cref{fig:Fl}) show the effects of our tested dampers on a legged system.
From touch-down to mid-stance (\textit{leg flexion}), the `free drop' curves show a larger negative work compared to the `slow drop' curves, illustrating that the damper absorbs extra energy. The returning curves (mid-stance to lift-off) of the hydraulic damper aligns well with the `slow drop' curve, indicating the damper is successfully detached due to slackness cable while the spring recoil. \Cref{fig:Fl_diaph} shows that the `free drop' force of the diaphragm damper is slightly higher than `slow drop' force in the first half of the leg extension phase. This discrepancy is likely caused by the elastic force component of the diaphragm damper due to sudden expansion of the air chamber volume. The elastic component seems to dominate the damper behavior, which thus acts mostly as an air spring. By separating its energetic components (\Cref{eqn10}), we found that the hydraulic damper produces a viscous-like resistance higher than the diaphragm damper (\SI{59}{mJ} versus \SI{2}{mJ}), indicating the hydraulic damper is more effective in dissipating energy under drop impact. Hence, the hydraulic damper shows more viscous behavior, while the diaphragm damper is more elastic.

Physical damping in the system comes at the cost of energy loss, and to maintain periodic hopping, it becomes necessary to replenish \mrkb{energy that is dissipated by damping ($E_{D_0}$)}. Therefore, there is a trade-off to consider: simulation results show that higher damping results in faster rejection of ground perturbation at the price of more energy consumption at reference drop height (\Cref{tab:energy}, \Cref{fig:energySimulation}). An adjustable damper would partly address this problem: on level ground, the damping rate could be minimal, and on rough terrain increased. The adjustability of the two dampers is illustrated in \Cref{fig:tunability_ind,fig:tunability_diaph}.  We discuss the adjustability from both energy dissipation and dynamic behavior perspectives.

Compared with the spring-only results, both the hydraulic and the diaphragm damper reduced the maximum leg flexion and dissipated more energy. The orifice setting changes the shape of the work loop differently for the two setups. For the hydraulic damper (\Cref{fig:tunability_ind}), orifice setting-c shrinks the work loop from left edge, indicating more resistance is introduced by the damper to reduce leg flexion. \mrka{For the diaphragm damper (\Cref{fig:tunability_diaph}), orifice setting-c not only shrinks the work loop, but also increases its slope. We interpret this as the elastic contribution of air compression: relatively fewer air leaves through the smaller orifice, but instead acts as an in-parallel spring.}

Concerning energy dissipation, changes of orifice settings led to relatively small changes in effective dissipated energy $E_\mathrm{effective}$: \SI{150}{mJ} to \SI{156}{mJ} for the hydraulic damper, and \SI{100}{mJ} to \SI{102}{mJ} for the diaphragm damper. Even for the other damper model (1210M), which dissipates high amounts of energy, changes in orifice setting change the work loop shape drastically, but not the dissipated energy (\SI{395}{mJ} versus \SI{401}{mJ}).

Similar to the isolated damper drop, the data (\Cref{fig:tunability_ind,fig:tunability_diaph}) shows that specific orifice settings introduce more resistance, but not necessarily lead to higher energy dissipation, for both hydraulic and diaphragm damper. \mrkb{However, in our simplified numerical leg model, an increase in viscous damping coefficients leads to a systematic increase of dissipated energy (\Cref{tab:energy}), and a sharper tip at the left side of the work loop (\Cref{fig:tunability_sim}). The discrepancy is likely due to the non-linear coupling between 
the damper mechanics and the leg dynamics in the hardware setup: (1) The damping force generated by the fluid dynamics in the orifice only approximates a linear viscosity \citep{Dixon2008}. (2) The impact loading on both the nonlinear leg structure and the damper. This makes the prediction of the energy dissipation not straight-forward based on our simplified numerical leg model, and points towards the need of a combined approach between simulation and hardware testing to fully understand physical damping in a legged system.} 

\mrkb{Viscous, velocity dependent damping alters the leg's loading characteristics, and leads to a peak force at the instance of touch-down. As a result, the vertical GRF is increased in the early stance phase, shifting and increasing the peak vertical GRF before mid-stance (\Cref{fig:time_GRF}). When designing a legged system with a viscous damper, its increasing load on the mechanical structure should be considered.}

\mrkb{The selection of viscous dampers depends on the task. High damping can fully reject disturbances in a single cycle, but lower damping could have energetic benefits. Here we looked for a damper that would dissipate significant negative work ($\frac{E_\mathrm{viscous}}{E_\mathrm{T_{0}}}\approx\SI{10}{\%}-\SI{15}{\%}$) in form of viscous damping. The air-filled diaphragm damper lead to insufficient energy losses (\SI{2}{\%}), but the hydraulic dampers dissipated \SI{10}{\%} and {60}{\%} of the system's total energy (\Cref{tab:energyComposition}).}

\mrkb{Drawing conclusions about animal locomotion based on the here presented leg-drop experiments is somewhat early. However, observations from \cite[Table 1, p.\,2288]{muller_kinetic_2014} indicate that leg forces can increase at unexpected step-downs during locomotion experiments. \marginpar{Rev1, GC 1} Further, \cite{Kalveram2012} suggests in a comparison of experimental human hopping and numerical simulations that damping may be \emph{the} driving ingredient in passive stabilization against ground-level perturbations. We are consequently excited about the here presented results of viscous dampers mounted in parallel to a leg's spring, producing adaptive forces without the need for sensing.}

\section{Conclusion}
We investigated the possibility to exploit physical damping in a simplified leg drop scenario as a template for the early stance phase of legged locomotion. Our results from a) numerical simulation promote the use of adjustable and viscous damping \mrkb{over Coulomb damping} to deal with a ground perturbation by physical damping. As such, we b) tested two technical solutions in hardware: a commercial, off-the-shelf hydraulic damper, and a custom-made, rolling diaphragm damper. We dissected the observed dissipated energy from the hardware damper-spring leg drops, into its components, by experimental design. The resulting data allowed us to characterize dissipation from the early impact (unsprung-mass effects), viscous damping, Coulomb damping, and orifice adjustments  \emph{individually, and qualitatively}. The rolling diaphragm damper features low-Coulomb friction, but dissipates only low amounts of energy through viscous damping. The off-the-shelf, leg-mounted hydraulic damper did exhibit high viscous damping, and qualitatively showed the expected relationship between impact speed, output force and negative work. Changes in orifice setting showed only minor changes in overall energy dissipation, but can lead to large changes in leg length dynamics, depending on the chosen technical damper. Hence, switching between different viscous, hydraulic dampers is an interesting future option. Our results show how viscous, hydraulic dampers react \mrkb{velocity-dependent}, and create an instantaneous, physically adaptive response to ground-level perturbations without sensory-input. 

\section*{Conflict of Interest Statement}
The authors declare that the research was conducted in the absence of any commercial or financial relationships that could be construed as a potential conflict of interest.

\section*{Author Contributions}
AM contributed to concept, hardware design, experimental setup, experimentation, data discussion and writing.
FI contributed to concept, simulation framework, experimental setup, data discussion and writing.
DH and ABS contributed to concept, data discussion and writing.

\section*{Acknowledgments}
The authors thank the International Max Planck Research School for Intelligent Systems (IMPRS-IS) for supporting AM, FI, the China Scholarship Council (CSC) for supporting AM, and the Ministry of Science, Research and the Arts Baden-W\"urttemberg (Az: 33-7533.-30-20/7/2) for supporting DH, and the Max Planck Society for supporting ABS.

\bibliographystyle{frontiersinSCNS_ENG_HUMS} \bibliography{literature}

\end{document}